# Automatic post-picking using MAPPOS improves particle image detection from Cryo-EM micrographs

Ramin Norousi[1,5], Stephan Wickles[2,5], Christoph Leidig[2], Thomas Becker[2], Volker J. Schmid[1], Roland Beckmann[2], Achim Tresch[3,4]

[1]Department of Statistics, Ludwig-Maximilians-University München, Germany

[2]Center for Integrated Protein Sciences and Munich Center for Advanced Photonics at the Gene Center, Department of Biochemistry, Ludwig-Maximilians-University München, Germany

[3]Max Planck Institute for Plant, Breeding Research, Cologne, Germany

[4]Institute for Genetics, University of Cologne, Cologne, Germany

[5]authors contributed equally to the work

Send correspondence to: tresch@mpipz.mpg.de, ramin@norousi.de, wickles@lmb.uni-muenchen.de

## ABSTRACT

Cryo-electron microscopy (cryo-EM) studies using single particle reconstruction are extensively used to reveal structural information on macromolecular complexes. Aiming at the highest achievable resolution, state of the art electron microscopes automatically acquire thousands of high-quality micrographs. Particles are detected on and boxed out from each micrograph using fully- or semi-automated approaches. However, the obtained particles still require laborious manual post-picking classification, which is one major bottleneck for single particle analysis of large datasets. We introduce MAPPOS, a supervised post-picking strategy for the classification of boxed particle images, as additional strategy adding to the already efficient automated particle picking routines. MAPPOS employs machine learning techniques to train a robust classifier from a small number of characteristic image features. In order to accurately quantify the performance of MAPPOS we used simulated particle and non-particle images. In addition, we verified our method by applying it to an experimental cryo-EM dataset and comparing the results to the manual classification of the same dataset. Comparisons between MAPPOS and manual post-picking classification by several human experts demonstrated that merely a few hundred sample images are sufficient for MAPPOS to classify an entire dataset with a human-like performance. MAPPOS was shown to greatly accelerate the throughput of large datasets by reducing the manual workload by orders of magnitude while maintaining a reliable identification of non-particle images.

## INTRODUCTION

The 3D cryo-electron microscopy (cryo-EM) reconstruction is a widely used and powerful technique for structural analysis of biological macromolecular assemblies. Due to the bad signal-to-noise-ratio (SNR) in low-dose cryo-EM, tens of thousands of 2D grayscale projections of a macromolecule (referred to as particles) are required to reconstruct its 3D structure (Woolford et al., 2007). The success of the reconstruction crucially depends on the number and the quality of 2D particle images. Non-particle images – i.e. ice, dust, contaminations, or empty images (noise) – in a dataset can lead to severe distortions in the result, including erroneous electron densities. Facing this problem, great efforts were made to accurately pick particles from the low-dose electron micrographs generated during cryo-EM. Excellent particle picking methods have been developed (Nicholson and Glaeser, 2001) and evaluated (Zhu et al., 2004). These methods can be divided into three categories according to the algorithms they use (Zhu et al., 2004): generative, discriminative, and unsupervised approaches. Generative approaches measure the similarity between regions of the micrograph and 2D projections of a reference structure. Most of these so-called template-matching techniques use cross-correlation as a similarity score (Chen and Grigorieff, 2007; Hall and Patwardhan, 2004; Huang and Penczek, 2004; Roseman, 2003). Instead of an initial reference structure discriminative methods require a training dataset, which contains positive and negative samples. According to these samples, a binary classifier is trained. This can be done either fully supervised using statistical learning (Hall and Patwardhan, 2004; Mallick et al., 2004; Volkmann, 2004) or machine learning (Arbelaez et al., 2011) approaches, or in an iterative, supervised fashion (Sorzano et al., 2009) allowing the user to correct the algorithm during the training phase. The category of unsupervised approaches entails algorithms that work without any reference. Particles are automatically detected based on statistical measures and features that are extracted directly from the micrograph (Adiga et al., 2005; Ogura and Sato, 2004; Roseman, 2003; Shaikh et al., 2008; Voss et al., 2009; Woolford et al., 2007; Zhu et al., 2004). All of these approaches focus on the optimization of correct particle detection on micrographs. These methods yield sets of boxed images whose quality depends on the signal-to-noise ratio of the micrograph, the picking method and the observed specimen. This can result in fractions of 10% to more than 25% of inadvertently selected particle images (Zhu et al., 2004). Although automated particle picking methods are invaluable for processing large cryo-EM datasets, subsequent manual curation is, therefore, still inevitable and now constitutes one major bottleneck for high resolution reconstruction of unsymmetrical particles.

Since the task of particle picking is distinct from the task of discriminating particles and non-particles in a collection of individual boxed images of standardized size we reasoned that both tasks should be addressed in individual steps. We propose to subject the output of automated particle picking methods to a specialized round of classification to separate particle images from non-particle images. To achieve this task we established MAPPOS (**M**achine learning **A**lgorithm for **P**article **POS**t-picking), a supervised discriminative post-picking method based on characteristic features calculated from the boxed images. Using MAPPOS, the manual curation step is reduced to the creation of a small training dataset that is then used to generate an ensemble of binary classifiers (Hansen and Salamon, 1990). Subsequently, this ensemble classifier can be applied to automatically sort a complete dataset. We used a simulation environment for the generation of realistic particle and non-particle images. With this controlled environment we had a tool at hand to accurately quantify the quality of post-picking classifications and used it to assess the performance of MAPPOS and compare it to the manual performances of experienced researchers (referred to as human experts).

## MATERIALS AND METHODS

### Simulation of a Cryo-EM Dataset

To develop and evaluate MAPPOS we required a synthetic yet realistic data environment. We generated 21,922 boxed images with a particle to non-particle ratio of 9:1, which is comparable to that of real cryo-EM datasets. The images were generated by projecting various 3D volumes into 2D. To resemble real cryo-EM images in fundamental properties – like their SNR and their image contrast modulated by the contrast transfer function (CTF) – all of the generated images were modified using a previously described image manipulation procedure (Baxter et al., 2009) (Fig. 1a). First, the structural noise present in real datasets was simulated by the addition of random noise with zero-mean Gaussian distribution to a SNR of 1.4. Second, the image formation of a bright field microscope at an acceleration voltage of 300kV and a defocus of 2.0 µm was simulated by modulation of the images with the corresponding contrast transfer function (CTF). Third, random noise (shot and digitization) of zero-mean Gaussian distribution was applied to a SNR of 0.05. In a final step, the simulated images were low-pass filtered to reduce the noise using a Butterworth filter between 20 and 50 Å. Low-pass filtering of micrographs or boxed images is routinely used during standard cryo-EM processing procedures to improve the visibility of the particles and, therefore, allow for better initial particle picking from micrographs or classification of boxed images (Shaikh et al., 2008). Particle images were projected from the crystal structure of the *E. coli* 70S ribosome (PDB: 2QAL & 2QAM (Borovinskaya et al., 2007)). Non-particle images were generated from four different 3D templates: plate, cylinder, sphere and void (Fig. 1b). These templates were chosen such that they covered the spectrum of contaminations typically encountered in cryo-EM datasets (Fig. 1c).

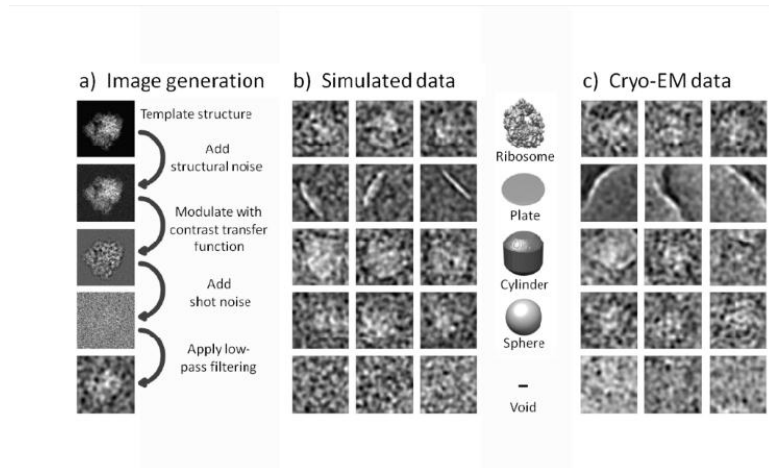

**Figure 1. Simulation of cryo-EM boxed images.** (**a**) Generation of an artificial cryo-EM image based on a crystal structure of the *E. coli* 70S ribosome (PDB: 2QAL & 2QAM). A 2D projection of the ribosomal electron density is modified by (i) addition of structural noise to account for structural heterogeneity, (ii) distortion with a CTF to simulate the image of a bright field electron microscope at a negative defocus value, (iii) addition of noise to a SNR of 0.05 to simulate low dose conditions, and (iv) low-pass filtering to improve the contrast as routinely done during standard cryo-EM image processing. (**b, c**) Comparison of experimental cryo-EM images (b) to our artificial projection images (c). Projections from different angles of the *E.coli* 70S ribosome (row 1) and four types of contaminations commonly found in cryo-EM datasets (rows 2-4) are shown. Various 3D volumes (middle column) were used to generate the artificial (non-)particle images.

**Testing of Discriminatory Features**

MAPPOS is based on the extraction of distinct features that enable discrimination of particle and non-particle images. Potential discriminatory features were tested individually on a dataset of *E. coli* 70S ribosomes that comprised 1,638 manually classified images with a 1:1 ratio of particles to non-particles. The discriminatory power of each feature was assessed by its receiver operating curve (ROC) and the associated area under curve (AUC), providing a balanced measure of its sensitivity and specificity (Bradley, 1997; Fawcett, 2006; Langlois and Frank, 2011).

**Description of Individual Discriminatory Features**

**Radially Weighted Average Intensity** (AUC-value = 0.83). The radially weighted average intensity is calculated as a weighted sum of the pixel intensities. The weighting is inversely proportional to a pixel's Euclidean distance from the image center. This measure for the centrality of bright pixels was found to assume higher values for particle images than for non-particle images.

**Phase Symmetry / Blob Detection** (AUC-value = 0.94). Blob detection is based on the concept of phase symmetry, a contrast- and rotation invariant measure of local symmetry at each point of an image. Phase symmetry recurs on a 2D Wavelet transformation that extracts local frequency information (Morlet et al., 1982). Phase symmetry transformation is applied with standard parameter settings as in (Kovesi, 1997). The transformed image is then binarized using Otsu's thresholding (Otsu, 1979). Subsequently, locally symmetric areas (blobs) can be counted. They were found to be mainly occurring in non-particle images (Fig. 2).

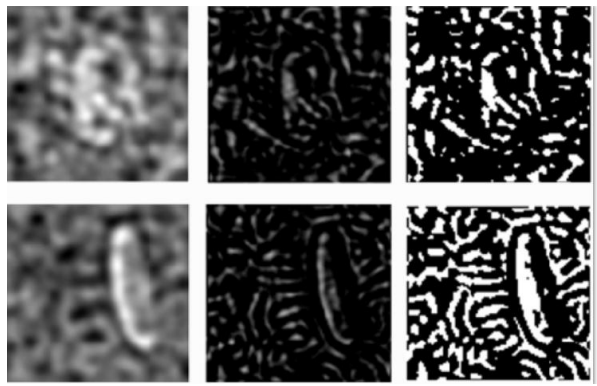

**Figure 2. Phase symmetry transformation and binarization.** A particle image (top row) and a non-particle image (bottom row) are depicted in their original states (left column), after phase symmetry transformation (middle

column), and after binarization (right column). The non-particle image contains an overall higher degree of symmetry and contains more white pixels after binarization.

**Dark Dot Dispersion** (AUC-value = 0.86). Images are convolved with a 2-dimensional symmetric Gaussian kernel. Afterwards they are binarized using the 95%-quantile of the image pixel intensity distribution as a threshold value. Connected regions of these pixels are referred to as dark dots. The center of a dark dot is determined as the mean of its pixel coordinates and the dark dot dispersion of an image is defined as the variance (mean squared Euclidean distance) between the centers of its dark dots. We observed that the average distance between dark dots is larger in particle images than in non-particle images.

**Creation of an Ensemble Classifier**

MAPPOS uses ensemble learning principles to construct ensemble classifiers from a set of individual classifiers. An ensemble classifier consists of a set of $k$ elementary independent classifiers $(C_1,…,C_k)$ for an identical classification problem. The $k$ binary predictions are combined to one final prediction by choosing the prediction made by the majority of the individual classifiers (Hansen and Salamon, 1990). An ensemble classifier generally yields an improved classification accuracy compared to each individual classifier (Duda et al., 2000). We implemented the bootstrap aggregating or 'bagging' approach (Breiman, 1996) for the construction of an ensemble. Prior to the learning procedure – to later assess the performance of the final ensemble classifier as unbiased as possible – a validation set comprising 10% of the training data is held aside. Once the final classifier is constructed, its performance is evaluated on the validation set. The $k$ elementary classifiers are iteratively selected out of a basic variety of classifiers and parameter settings. To that end, the remaining 90% of the training data are randomly split 5 times by subsampling an inner training set (80%) and an evaluation set (20%). We start with a set of (40 – 60) candidate classifiers selected from a variety of basic methods such as linear discriminant analysis (Mika et al., 1999), decision trees (Quinlan, 1986), support vector machines (Vapnik, 1995), and n-nearest-neighbors (Cover and Hart, 1967). Each candidate classifier is equipped with parameters randomly drawn from an appropriate range. The candidate classifiers are trained 5 times using the 5 inner training sets, respectively. Subsequently, they are applied to the corresponding 5 evaluation sets, and the classifier which performs best is added to the classifier ensemble. This process is iterated until no improvements can be made by the addition of another classifier (Wichard, 2006). In our case, the final classification ensemble contained 21 elementary classifiers.

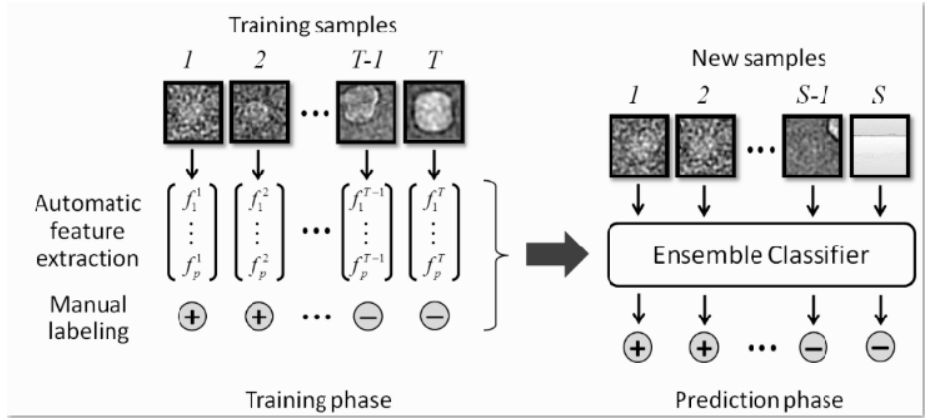

**Figure 3. Workflow of MAPPOS.** A training dataset is created by manual classification of $T$ sample images (typically, $T \approx 1,000$) as particles or non-particles labeled (+) or (-), respectively. During the training phase, $p$ discriminatory features $(f_1^j,…,f_p^j)$ are automatically extracted from each sample image $j$. The feature matrix $(f_k^j)$ is used in combination with the sample labels to train an ensemble classifier. The ensemble classifier is used during the prediction phase to efficiently classify all of the $S$ images (typically, $S \approx 10^5\text{-}10^6 \gg T$) of the complete dataset.

## Data Acquisition

*E. coli* 70S ribosomes were prepared as previously described in (Burma et al., 1985). In short, a crude ribosomal fraction was further purified using a liner 10-40% sucrose gradient. The monosomal fraction was then applied to carbon-coated holey grids. Micrographs were recorded under low-dose conditions (about 20 e$^-$ per Å$^2$) on a Titan Krios TEM (FEI Company) microscope at 300 kV at a magnification of 128,200× using an Eagle 4k × 4k CCD camera (FEI Company, 4,096 × 4,096 pixel, 15 µm pixel, 5 s/full frame) in a negative defocus range of 1.0–3.5 µm. The resulting pixel size was 1.17 Å on the object scale.

Data were collected using the semi-automated software EM-TOOLS (TVIPS GmbH). This allowed manual selection of appropriate grid meshes and holes in the holey carbon film. During acquisition, the software automatically performed a re-centering, drift and focus correction before the final spot scan series were taken. Long-term TEM instabilities in beam shift, astigmatism and coma were corrected by EM-TOOLS in regular intervals (for example, every 45 min).

## Validation

The performance of MAPPOS was evaluated on simulated and experimental datasets featuring *E. coli* ribosomes as particles. The experimental cryo-EM data comprised 85,726 boxed images of empty *E. coli* 70S ribosomes which were automatically picked by Signature (Chen and Grigorieff, 2007) using a template matching algorithm. The artificial datasets were obtained from simulated boxed images. Each of them comprised a set of 19,874 particle images and 2,048 non-particle images. The particles were randomly selected from the projections of the *E. coli* ribosome. We analyzed five scenarios with different non-particle sets. We either exclusively used plate, cylinder, sphere or void projections (see Fig.1), or a uniform mixture of projections of all these types.

For the simulated dataset the images were unambiguously known to contain particles or non-particles, respectively. For the experimental dataset the correct labels were assigned manually. A set of 2,000 images (particle to non-particle ratio of 1:1) was used as a training set for automated classification by MAPPOS, respectively. We compared three methods, (i) manual post-picking, (ii) no post processing and (iii) post-picking with MAPPOS. Standard performance measures were calculated for 2x2 contingency tables of true vs. predicted labels (Langlois and Frank, 2011). In addition, we assessed the quality of the resulting electron density map for the experimental dataset. Besides classification performance, we assessed the effects of the different post-picking results on the quality of the resulting electron density maps. All datasets were processed using SPIDER (Frank et al., 1996) and refined for 3 rounds to final resolutions of about 11 Å corresponding to the Fourier-shell correlation = 0.5 (FSC0.5) criterion.

## RESULTS

### MAPPOS

MAPPOS is a simple yet effective tool that has been implemented in MATLAB (version 6.0 or greater). It requires the image processing toolbox and the statistics toolbox as add-on packages. The source code for MAPPOS is freely available for academic use. It can be downloaded from http://www.treschgroup.de/mappos.html. A detailed step-by-step instruction for MAPPOS is provided in the Supplementary Material.

### MAPPOS Relies on Seven Features for Image Classification

We set out to develop a fast and reliable classification method for post-picking of boxed cryo-EM images into particles and non-particles. To achieve a high robustness we avoid any user-adjustable parameters, thereby minimizing the risk of over-fitting. We tested a number of discriminatory features for applicability to this problem and identified 7 well-performing features that constitute the input to MAPPOS: (i) location and scale (mean, variance), (ii) the (0%, 10%, 50% ,90% ,100%)-quantiles of the pixel intensity distribution, (iii) the number of foreground pixels after binarisation using Otsu's thresholding (Otsu, 1979), (iv) the number of edges counted after Canny edge detection (Canny, 1986) (see Supplementary Material), (v) the radially weighted average intensity, (vi) phase symmetry / blob detection, and (vii) dark dot dispersion. Discriminatory features (v) to (vii) are described in more detail in the Materials and Methods section.

Classification by MAPPOS is dependent on the availability of a relatively small set of sample images (training dataset) that have been labeled as particles (+) or non-particles (-) and can be divided into a learning phase and a particle detection (prediction) phase (Fig. 3). During the learning phase MAPPOS uses the training dataset to create an ensemble classifier from a pool of elementary classifiers (see Materials and Methods). This ensemble classifier is then used during the particle detection phase to classify an entire dataset.

## MAPPOS Classifies Boxed Images with High Speed

The learning of the ensemble classifier from a training dataset containing 2,000 images takes about 75 seconds on a standard PC (Intel Core Duo 2x2.00 GHz; 2 GB RAM). During the detection phase, images are classified with a speed of 0.3 seconds per image or 12,000 images per hour.

## Performance in a Simulated Data Environment

Non-particle contaminations in cryo-EM datasets can severely impair the quality electron density map reconstructions. Taking into account that contemporary automated data collection approaches typically provide an excess of raw data, we focused on a high detection rate for non-particles during the development of MAPPOS – at the expense of also removing some particles along the way. In terms of quantifiable measures we were trying to maximize the positive predictive value (PPV) and the specificity (Langlois and Frank, 2011) while accepting a lower sensitivity in return (see Table 1 for definitions).

Each of the five simulation scenarios (see Methods) was run 100 times, and each run provided a vector (TP, FP, TN, FN) of true positives (TP), false positives (FP), true negatives (TN), and false negatives (FN) as its result. The specificity and sensitivity (as well as their mean and variance values) for each scenario were derived from these values (Table 1). Specificity values were above 70% in all scenarios, reaching a maximal value of 86% for the cylinder scenario and a value of 74% for the mixed scenario. Sensitivity values were around 80% with the exception of the cylinder scenario with a value of 87%.

| Conta-mination type | TP ±relative std.dev. | FP ±relative std.dev. | TN ±relative std.dev. | FN ±relative std.dev. | Sensitivity ±std.dev in % points | Specificity ±std.dev in % points |
|---|---|---|---|---|---|---|
| Plate | 814 ±2% | 300 ±6% | 701 ±2% | 186 ±10% | 81% ±2% | 70% ±2% |
| Cylinder | 865 ±1% | 144 ±10% | 856 ±2% | 201 ±9% | 87% ±1% | 86% ±1% |
| Sphere | 798 ±2% | 241 ±8% | 759 ±3% | 202 ±9% | 80% ±2% | 76% ±2% |
| Void | 806 ±3% | 296 ±7% | 704 ±3% | 194 ±12% | 81% ±2% | 70% ±2% |
| All | 794 ±2% | 257 ±7% | 743 ±2% | 206 ±9% | 79% ±2% | 74% ±2% |

**Table 1. Performance of MAPPOS in different test scenarios.** A set of 1,000 particles and 1,000 non-particles was used for training in each case. For self-containedness, we provide a definition of the performance scores proposed by (Langlois and Frank, 2011) for the comparison of particle picking methods. Classification results on the test set were compared to the known labels, splitting the samples into correctly classified particles (true positives, TP), correctly classified non-particles (true negatives, TN), incorrectly classified non-particles (false positives, FP), and incorrectly classified particles (false negatives, FN). Quantities that are derived from these numbers are the sensitivity (=TP/(TP+FN)), specificity (=TN/(TN+FP)), and positive predictive value (PPV=TP/(TP+FP)).

**MAPPOS Performs Comparable to Human Experts**

MAPPOS was compared to 7 human experts using a simulated dataset of 2,048 images comprising 1,638 particles and 410 mixed non-particles. We trained MAPPOS with two different training datasets. The first one comprised 500 true particles and 500 true non-particles, while the second one comprised 500 particles and 500 non-particles that were randomly chosen from the classified particles of the best-performing human expert. The results were analyzed according to sensitivity and specificity (Fig. 4). Notably, the performances of the human experts were highly variable. The results achieved by MAPPOS were comparable to those of the best-performing human experts. When trained with the first training dataset, MAPPOS scored the 2$^{nd}$ best specificity, achieving 79% specificity and 81% sensitivity. When trained with the random images obtained from the human expert, the specificity (67%) was still above average, while the sensitivity (85%) increased considerably. We analyzed the individual classified particles according to the types of non-particles that were detected with high or low reliability, respectively. MAPPOS agreed best with the two most accurate (specificity) human experts, and, when MAPPOS was trained by a human expert, it mimicked this expert's classification behavior (Fig. 5 and data not shown).

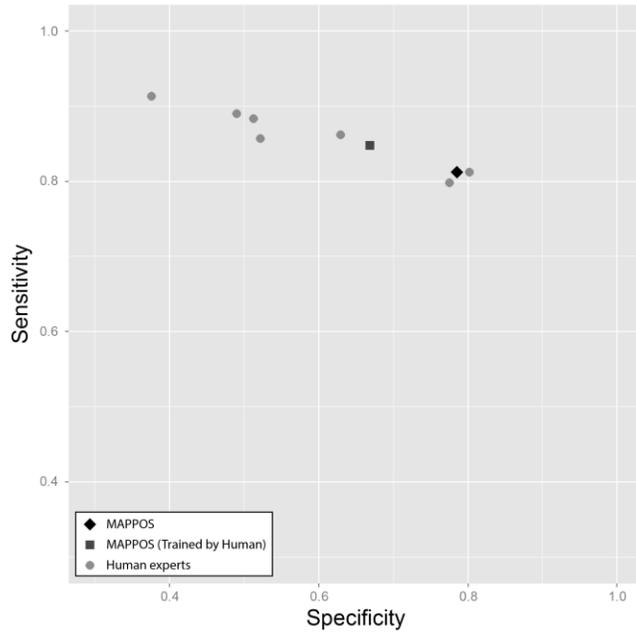

**Figure 4. Comparison between MAPPOS and human experts.** Sensitivity and specificity values of 7 human experts (circles), MAPPOS trained by the best-performing human expert (square) and MAPPOS trained with true particles and non-particles (rhombus) are shown.

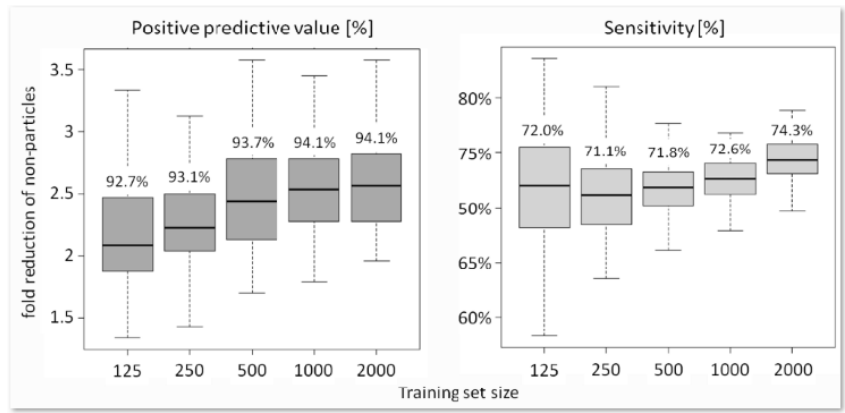

**Figure 5. Effect of training dataset size on MAPPOS performance.** The PPV (left) and the sensitivity (right) of MAPPOS were tested on experimental cryo-EM data of the *E. coli* 70S ribosome sample using various training dataset sizes (x-axis). Each box summarizes the results of 100 replicate *in silico* experiments. Each box spans the central range of the data (1$^{st}$-3$^{rd}$ quartile) while the black lines inside the boxes indicate the respective medians. The whiskers mark the 3-fold inter-quartile range. The fold reduction of the number of non-particles in the dataset (y-axis) is indicated for the PPV.

**Performance with Experimental Cryo-EM Data**

Particles were picked from micrographs containing *E. coli* 70S ribosomes using Signature (Chen and Grigorieff, 2007) and subsequently classified by MAPPOS. The same particles were manually inspected and classified by a human expert. First, we investigated the PPV and sensitivity as a function of the training dataset size (Fig. 5). Both, sensitivity and PPV increased with the size of the training dataset. The maximal PPV was already achieved with a training dataset of only 1,000 images. The increase of the PPV from 86% (PPV of Signature with manual classification as the reference) to 94% after post-picking by MAPPOS corresponds to a substantial reduction of the fraction of false positives by a factor of about 2.5. Second, as an additional measure of performance, we evaluated the quality of the electron density maps reconstructed from the MAPPOS or human expert datasets in terms of structural features that were clearly resolved. We used the backprojection of the crystal structure of the *E. coli* 70S ribosome and a reconstruction based on the unclassified dataset as a positive and a negative control, respectively. According to Fourier-Shell correlation the resolution of the reconstructions was comparable in all cases; however, there were obvious differences in the quality of the density maps. Considering the ribosome, the and, later on, β-sheets. One of the evaluated regions was the ribosomal tunnel exit at the ribosomal proteins L29 and L23 (Fig. 6). No separation between rRNA and protein densities was observed in the negative control. Accordingly, no information on protein secondary structure information could be obtained. The reconstructions of the dataset classified by MAPPOS as well as the manually inspected dataset provide information on the localization and secondary structure of proteins. The α-helices of L29 and L23 are almost completely resolved. Our results illustrate how post-picking of automatically selected particles from cryo-EM micrographs can lead to improved electron density maps, and that post-picking by MAPPOS is on par with manual particle inspection.

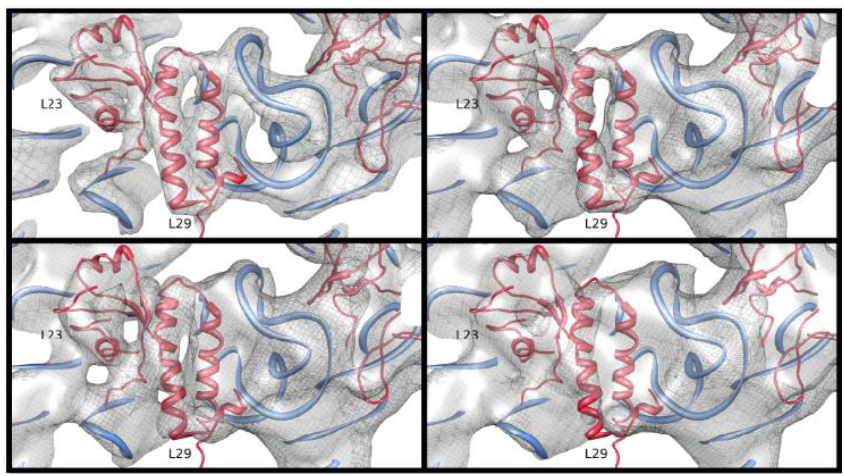

**Figure 6. Effect of different post-picking classification strategies on cryo-EM reconstructions.** A ribbon representation (red: ribosomal proteins; blue: rRNA) of a crystal structure of the *E.coli* 70S ribosome (PDB: 2QAL & 2QAM) was fitted in all electron density maps (grey). The electron density map projected from the crystal structure of the *E. coli* 70S ribosome and filtered to 10 Å resolution serves as a reference (top left). Secondary structure elements, e.g. protein α-helices of L29, are resolved in the reconstruction of the manually classified dataset (bottom left) as well as in the reconstruction of the dataset classified by MAPPOS (top right). In contrast, no secondary structure information is resolved in the reconstruction of the unclassified dataset (bottom right).

## DISCUSSION

We introduced MAPPOS, an automated image classification method that reduces the manual workload of particle post-picking by orders of magnitude while maintaining a reliable classification quality. We used the example of the *E. coli* ribosome to demonstrate that the quality of the electron density map after reconstruction from an automatically classified dataset is equal to that of a manually classified dataset.

When compared to human experts, MAPPOS achieved a performance similar to the best-performing human experts. Notably, the performances of the human experts varied considerably (Fig. 5). An obvious concern resulting from this observation is that MAPPOS cannot perform better than the user that classified the training dataset. However, in the light of large datasets, it is easier for a user to thoroughly assemble a good training dataset than to uniformly judge all of the boxed images with constant high quality. It is exactly this task where MAPPOS outperforms manual classification by impartially applying the same criteria to every image even throughout large datasets.

Existing methods (Nicholson and Glaeser, 2001) for automated particle selection aim at the simultaneous identification and classification of particles on the level of micrographs. The crucial difference in our approach is to address these tasks separately. We see the major advantage of this strategy in the possibility to provide positive (particle) as well as negative (non-particle) samples that were derived directly from the dataset itself for the training of the classifier. Such sample images are not available prior to particle picking; only samples from previous datasets or unrelated references can be provided. The use of negative samples that resemble the actual types of non-particles present in a dataset contributes greatly to the specificity achieved by MAPPOS. Consequently, the preceding particle picking step can be highly sensitive (filter criteria can be less strict), because a sufficient specificity is ensured by the subsequent post-picking step.

In this study, we demonstrated the applicability of MAPPOS to experimental cryo-EM data on the example of an *E. coli* 70S ribosome dataset. While the application of MAPPOS is probably most beneficial for unsymmetrical specimen since large numbers of particles are required for cryo-EM reconstructions in such cases, it remains to be shown that it is applicable to a broader range of particles. Notably, MAPPOS has already been successfully used for high-resolution reconstructions of 80S ribosomal complexes (Becker et al., 2012; Leidig et al., 2012).

Current generation electron microscopes (e.g. FEI Titan Krios) generate up to 4,000 micrographs per day using automated data acquisition techniques (Otto Berninghausen & Ingo Daberkov, personal communication). For ribosomal samples this amounts to roughly 200,000 – 500,000 particles per day. Automated particle picking and post-picking tools are therefore likely to become an integral part of cryo-EM processing pipelines. Manual classification of the *E. coli* 70S ribosome dataset used in this study required 3 – 4 working days although it comprised merely 85,726 boxed images. In contrast, the classification using MAPPOS required less than a day – including the manual generation of the training dataset. Despite its high speed, the quality of the final 3D reconstruction was equivalent to that of the manually classified dataset. This demonstrates that MAPPOS is able to handle huge amounts of data at the same pace at which they are generated without impairing the quality of the resulting electron density maps.

## AUTHORS' CONTRIBUTIONS

Achim Tresch and Roland Beckmann initiated the research. Ramin Norousi, Volker Schmid, and Achim Tresch developed the algorithm. Stephan Wickles, Ramin Norousi, Thomas Becker, Christoph Leidig, Roland Beckmann and Achim Tresch analyzed the data. Achim Tresch, Ramin Norousi, Christoph Leidig and Stephan Wickles wrote the manuscript. All authors read and approved the final version of the manuscript.

## ACKNOWLEDGEMENTS

We thank the 7 human experts Thomas Becker, Julian Deeng, André Heuer, Christoph Leidig, Eli van der Sluis, Marco Wachowius and Stephan Wickles for their participation in the study and valuable comments on the manuscript. This work was supported by the Deutsche Forschungsgemeinschaft (SFB646). Achim Tresch was supported by a Gastprofessur grant from the Deutsche Forschungsgemeinschaft to Patrick Cramer.